\documentclass[letterpaper]{article} 
\usepackage{aaai2026}
\usepackage{times}  
\usepackage{helvet}  
\usepackage{courier}  
\usepackage[hyphens]{url}  
\usepackage{graphicx} 
\usepackage{natbib}  
\usepackage{caption} 
\usepackage{algorithm}
\usepackage{algorithmic}

\usepackage{amsmath}
\usepackage{amssymb}   

\usepackage{multirow}   

\usepackage{booktabs}
\usepackage{tablefootnote}

\usepackage{newfloat}
\usepackage{listings}
\DeclareCaptionStyle{ruled}{labelfont=normalfont,labelsep=colon,strut=off} 
\floatstyle{ruled}
\newfloat{listing}{tb}{lst}{}
\floatname{listing}{Listing}
\title{Feature-Aware One-Shot Federated Learning via Hierarchical Token Sequences}

\author {
    Shudong Liu\textsuperscript{\rm 1},
    Hanwen Zhang\textsuperscript{\rm 1},
    Xiuling Wang\textsuperscript{\rm 2},
    Yuesheng Zhu\textsuperscript{\rm 1},
    Guibo Luo\textsuperscript{\rm 1}\thanks{Corresponding author}
}

\affiliations {
    \textsuperscript{\rm 1}Guangdong Provincial Key Laboratory of Ultra High Definition Immersive Media Technology,\\
    Shenzhen Graduate School, Peking University\\
    \textsuperscript{\rm 2}Hong Kong Baptist University\\
    \{shooo, hanwen\}@stu.pku.edu.cn, \{zhuys, luogb\}@pku.edu.cn, xiulingwang@hkbu.edu.hk

}

\usepackage{bibentry}

\begin{document}

\maketitle

\begin{abstract}
One-shot federated learning (OSFL) reduces the communication cost and privacy risks of iterative federated learning by constructing a global model with a single round of communication. However, most existing methods struggle to achieve robust performance on real-world domains such as medical imaging, or are inefficient when handling non-IID (Independent and Identically Distributed) data. To address these limitations, we introduce FALCON, a framework that enhances the effectiveness of OSFL over non-IID image data. The core idea of FALCON is to leverage the feature-aware hierarchical token sequences generation and knowledge distillation into OSFL. First, each client leverages a pretrained visual encoder with hierarchical scale encoding to compress images into hierarchical token sequences, which capture multi-scale semantics. Second, a multi-scale autoregressive transformer generator is used to model the distribution of these token sequences and generate the synthetic sequences. Third, clients upload the synthetic sequences along with the local classifier trained on the real token sequences to the server. Finally, the server incorporates knowledge distillation into global training to reduce reliance on precise distribution modeling. Experiments on medical and natural image datasets validate the effectiveness of FALCON in diverse non-IID scenarios, outperforming the best OSFL baselines by 9.58\% in average accuracy. 
\end{abstract}

\begin{links}
\link{Code}{https://github.com/LMIAPC/FALCON}
\end{links}


\section{Introduction}

Federated Learning (FL) is a distributed collaborative training paradigm that enables the integration of data from multiple parties without exposing raw data~\cite{avg17}. However, conventional iterative FL frameworks require multiple rounds of parameter synchronization between clients and the central server, leading to significant communication overhead. Moreover, frequent interactions increase the risk of privacy leakage~\cite{fladvance21} and make the system more vulnerable to poisoning \cite{24poison, 25poison} or inference attacks~\cite{nasr2019comprehensive, lyu2020threats}. To alleviate these limitations, one-shot federated learning (OSFL) frameworks have been proposed~\cite{guha2019}, where each client uploads its local information only once, substantially reducing communication cost and mitigating privacy risks.



While OSFL models show promise in FL, their practical deployment is hindered by non-IID data distributions across clients, while insufficient efficiency in global modeling limits large-scale deployment.

Existing OSFL methods can be broadly classified into two categories: those that upload local model parameters and those that transmit surrogate data in place of original data to facilitate global model construction. 

Among parameter-uploading methods, some studies focus on designing weight aggregation strategies \cite{fusefl24, fisher24}. However, these methods typically assume strong homogeneity among local models; the distributional shifts often lead to divergence in local models, resulting in suboptimal global performance. Bayesian ensemble approaches \cite{fedbe21, bpb24, fens24} improve robustness to non-IID data by modeling and fusing the local posteriors. However, they typically incur high inference costs and require access to public data for posterior estimation. Knowledge distillation–based methods \cite{feddf20, dense22, isca23, coboosting24} transfer knowledge from multiple local models to a global model using auxiliary data, such as public datasets or synthetic samples. Although they relax model homogeneity constraints, their effectiveness depends heavily on high-quality auxiliary data, limiting practical deployment and overall performance.

\begin{figure}[t]
\centering
\includegraphics[width=1\columnwidth]{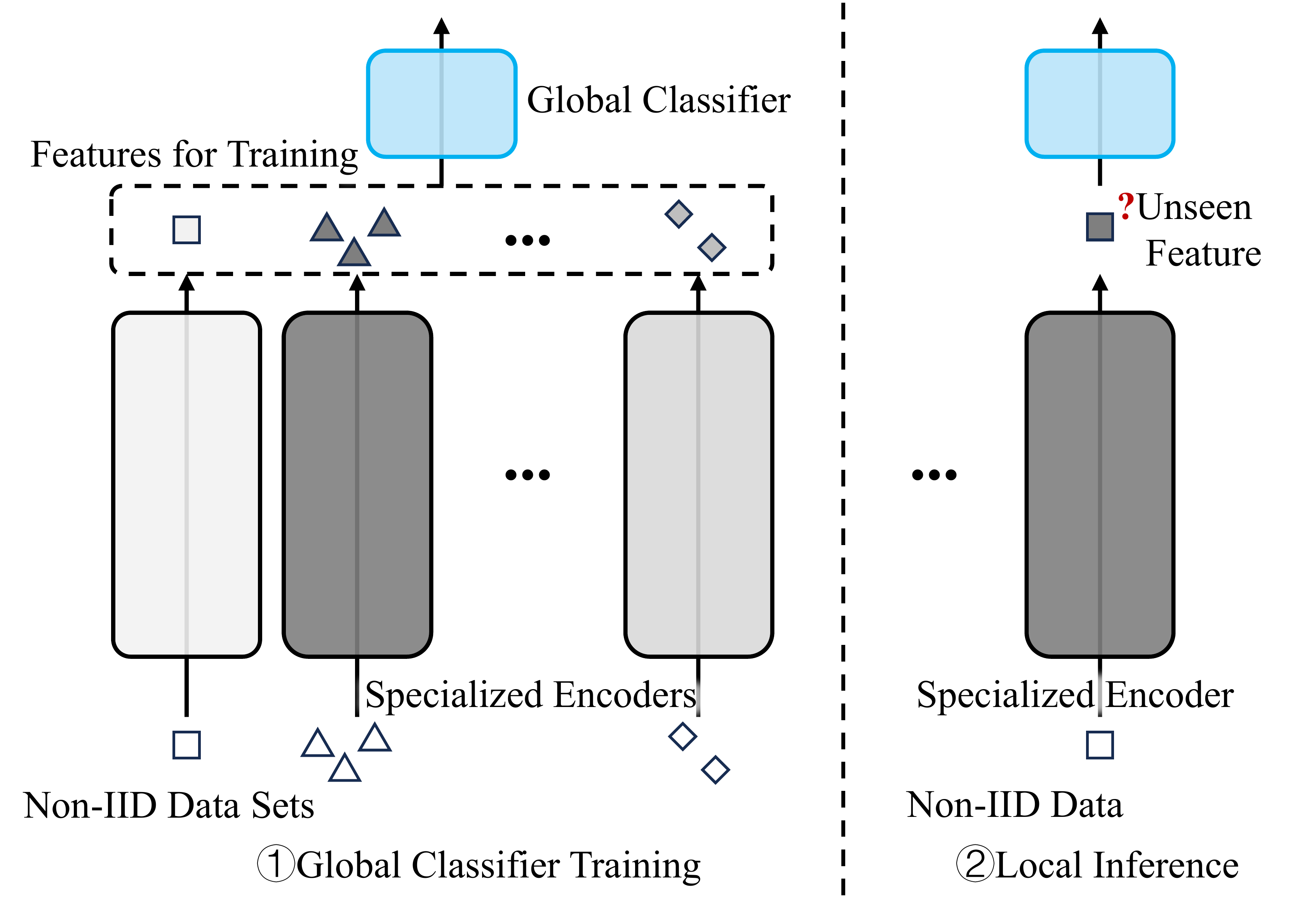}  
\vspace{-0.1in}
\caption{Illustration of the impact of locally specialized encoders on generalization. 
Colors indicate different locally specialized encoders; shapes indicate samples from non-IID datasets.
(1) Global classifier training: Features from locally specialized encoders are aggregated to train a global classifier.
(2) Local inference: When a client's specialized encoder encounters data that is non-IID with respect to its local data, the extracted features cannot be correctly recognized by the global classifier.}
\label{fig:spec_enc}
\vspace{-0.1in}
\end{figure}

Surrogate data uploading methods aim to mitigate the performance degradation caused by non-IID client distributions. 
Dataset distillation approaches \cite{dosfl20, fedd323, sd2c24} transmit a small amount of distilled data, which often fails to represent the full complexity of the original data distribution. 
Recently, distribution-estimation–based surrogate data methods have demonstrated advantages in both performance and efficiency. These methods primarily fall into two categories: (i) image-level generative-model–based approaches, and (ii) feature distribution statistical modeling approaches. 

Image-level generative-model–based approaches \cite{osgan23, fedcvae23, fgl23, feddeo24, fedbip25, lmg25} aim to reconstruct the client-side data distribution on the server using GANs \cite{gan20}, VAEs \cite{vae13}, or diffusion models \cite{ddpm20, ldm22}, thereby improving the global model’s adaptability to non-IID data. However, GANs are prone to mode collapse and unstable training \cite{wgan17}, while VAEs often produce blurry samples. Pretrained diffusion models struggle to produce usable samples in real-world application domains such as medical imaging. Moreover, the reliance on iterative denoising steps is a key bottleneck: even with fast solvers\cite{ddim, pndm}, generating training samples on the server remains time‑consuming, hindering practical deployment. 

Feature distribution statistical modeling approaches use a unified encoder to extract semantic features from raw data, then perform statistical modeling on the features. By uploading the distribution statistics, the server can reconstruct a global feature space. Specifically, FedPFT~\cite{pft} models features as a Gaussian mixture, uploading estimated statistics for pseudo-sample generation. FedCGS\cite{cgs25} aggregates global feature statistics and directly builds a Gaussian naive Bayes classifier without further training. However, features extracted by directly applying a pretrained model often perform suboptimally in real-world tasks with distribution shifts from the encoder’s pretraining domain. While local fine-tuning is a natural way to obtain a specialized encoder, it is not feasible in OSFL. Figure~\ref{fig:spec_enc} illustrates a simplified semantic feature transmission OSFL scenario. Local specialization leads to misaligned feature spaces across clients, resulting in personalized local models rather than a true global model—thus failing to meet the generalization requirement of FL. Moreover, most existing pretrained visual encoders based on Vision Transformers \cite{attn17, vit} accept fixed-size low-resolution inputs, which limits their ability to capture fine-grained regional details and local structures, particularly critical in high-resolution tasks such as medical imaging. Thus, with the restriction that the encoder cannot be locally fine-tuned, a key challenge is enhancing the adaptability of pretrained encoders to multi-domain data while efficiently extracting multi-scale and high-resolution information from raw data. Finally, the effectiveness of distribution-estimation–based methods is constrained by the challenge of accurately replicating client data distributions, which is rarely attainable in practice.

To overcome the above limitations, we propose FALCON, a one-shot federated learning framework designed for diverse application domains and multi-type non-IID scenarios. 
Specifically, FALCON employs a pretrained visual encoder on each client and introduces the proposed hierarchical scale encoding (HSE) to simultaneously extract both global semantic features and high-resolution regional features from images. This process constructs hierarchical token sequences with multi-scale semantics. Each client subsequently trains a local classifier and a multi-scale autoregressive (M-AR) transformer generator to capture the distribution of token sequences and generate the synthetic token sequences. The server receives the synthetic sequences and local classifiers from clients and applies knowledge distillation to complement generative modeling and enhance global training. Our contributions are as follows:

\begin{itemize}
    \item 
    We propose FALCON, a novel OSFL framework that achieves efficient global modeling on non-IID data by uploading designed compressed surrogate data and leveraging distillation-guided optimization.
    \item 
    We propose HSE to extract multi-scale semantics from raw images for better domain adaptability, and design an M-AR generator to model hierarchical token sequences while preserving spatial relationships.
    \item We incorporate a distillation-guided global training mechanism that reduces the reliance on synthetic sequence quality and facilitates robust global modeling. 
    \item Experiments on natural and medical image datasets under non-IID settings demonstrate the effectiveness, efficiency, and privacy-preserving capability of FALCON.
\end{itemize}

\section{Method}
\subsection{Problem Formulation}
We consider a federated learning (FL) setting consisting of a central server and \( N \) distributed clients. Each client \( n \) holds a private local dataset 
$\mathcal{D}_n = \left\{ (x_i, y_i) \right\}_{i=1}^{|\mathcal{D}_n|}$
.
The server aims to train a global model \( w(\cdot) \) based on information uploaded by the clients, without accessing the original data, so as to minimize the sum of empirical risks across all clients. The FL optimization objective can be formulated as:
\begin{equation}
\min_{w} \mathcal{L}(w)
= \frac{1}{\sum_{j=1}^{N}\lvert \mathcal{D}_j\rvert}
  \sum_{n=1}^{N}
  \sum_{(x, y)\in\mathcal{D}_n} \bigl[\ell\bigl(w(x),y\bigr)\bigr],
\label{eq:fl-objective}
\end{equation}
where \( \ell(\cdot, \cdot) \) denotes the task-specific loss function.


\begin{figure*}[t]
\centering
\includegraphics[width=0.95\textwidth]{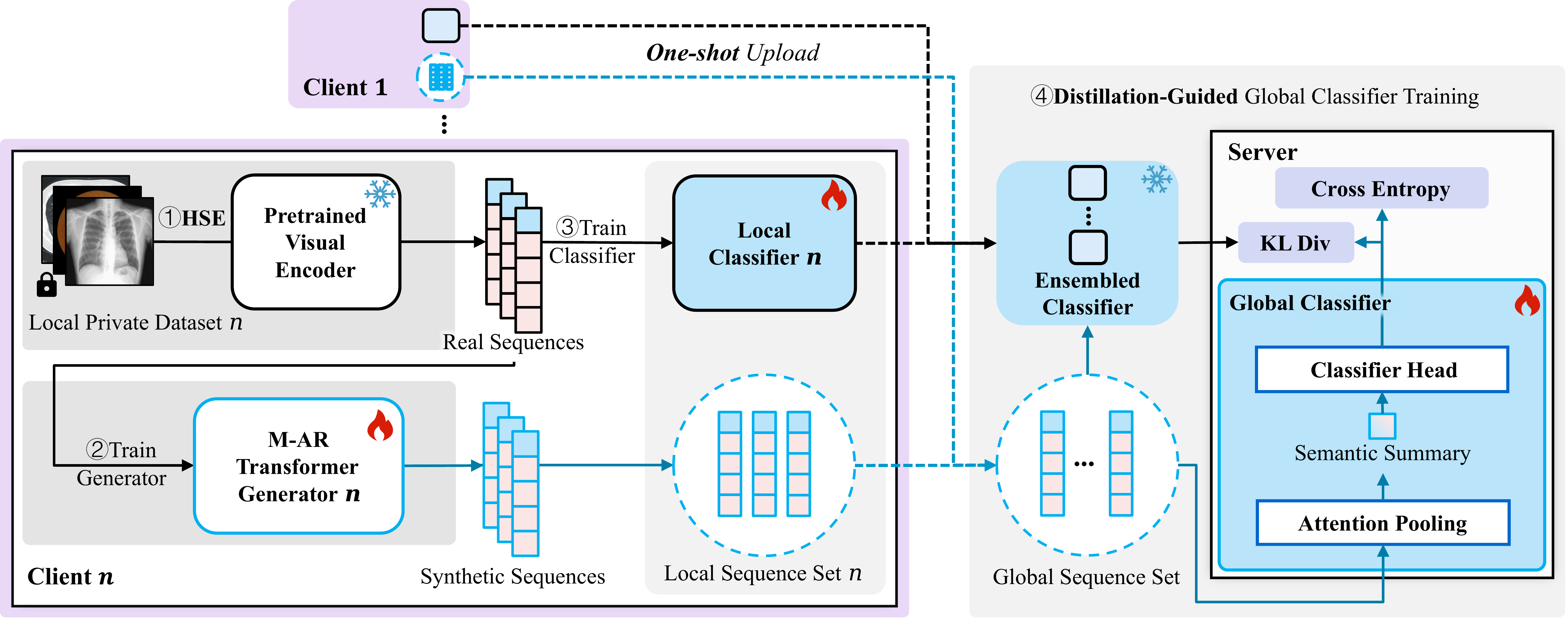}
\caption{The framework of FALCON. 
Each client (1) performs hierarchical scale encoding (HSE), (2) trains a local M-AR transformer generator (3) and classifier, and uploads synthetic token sequences and the classifier to the server for (4) distillation-guided global classifier training.
}
\label{fig:fm}
\vspace{-0.1in}
\end{figure*}

\subsection{FALCON Framework}

FALCON enables one-shot federated learning through the transmission of compact, structured token sequences and distillation-guided training of client knowledge.
As shown in Figure~\ref{fig:fm}, FALCON proceeds in four steps:
(1) Each client compresses its raw images into hierarchical token sequences via hierarchical scale encoding (HSE);
(2) The client trains a multi-scale autoregressive (M-AR) generator on these sequences;
(3) It also trains a local classifier. 
Both the synthetic token sequences (sampled from the generator) and the classifier are then uploaded to the server;
(4) Finally, the server performs distillation-guided training using the uploaded sequences and local classifiers to optimize the global model.

\subsection{Hierarchical Scale Encoding (HSE)}

\begin{figure}[t]
\centering
\includegraphics[width=0.88\columnwidth]{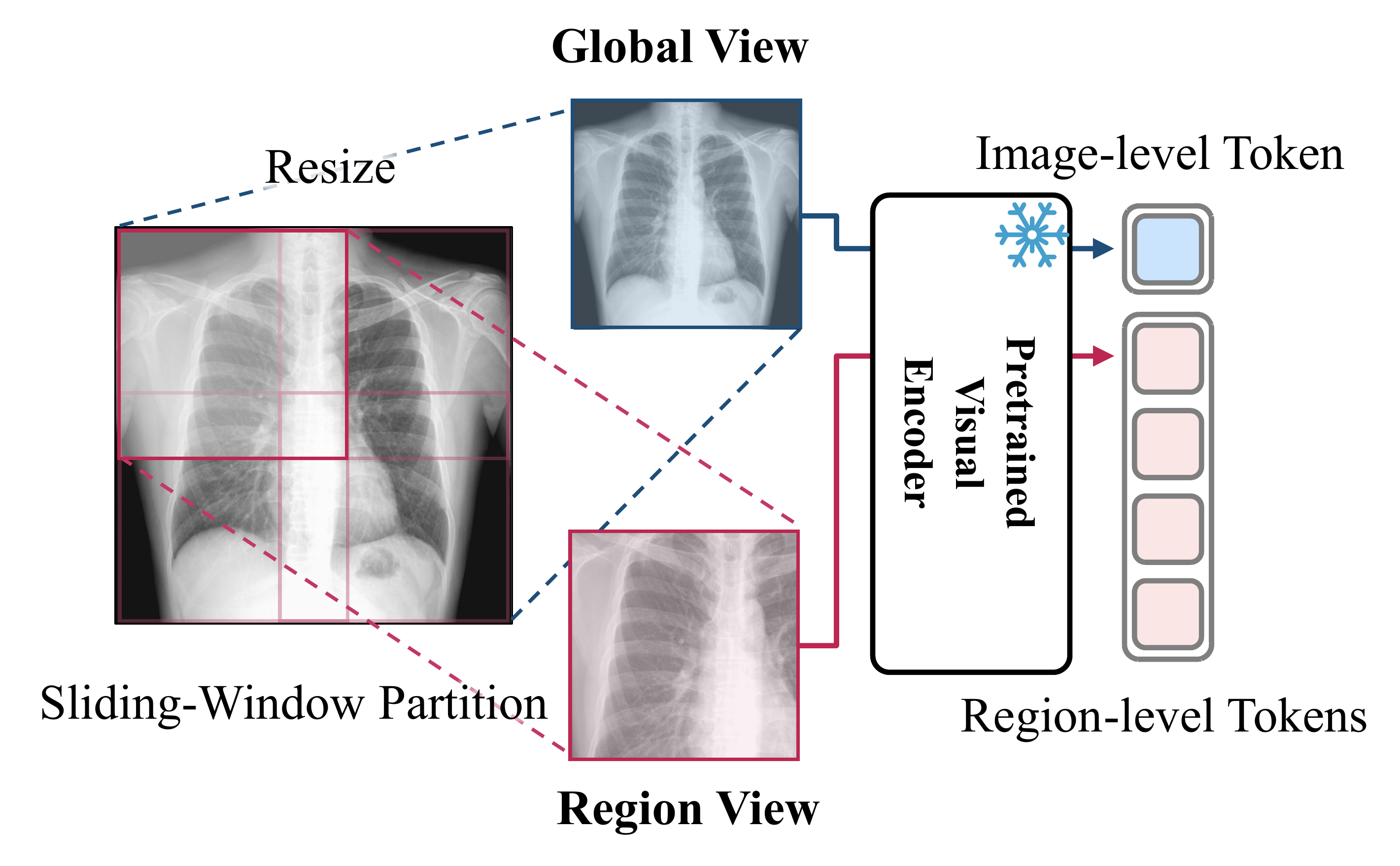}  
\vspace{-0.1in}
\caption{Illustration of hierarchical scale encoding (HSE).}
\label{fig:hse}
\vspace{-0.1in}
\end{figure}

To capture both global discriminative information and high-resolution local details without locally fine-tuning the pretrained visual encoder, FALCON draws inspiration from the human visual system, which first perceives the global scene and then attends to local regions. Accordingly, we design hierarchical scale encoding (HSE). 
As shown in Figure~\ref{fig:hse}, we employ two complementary encoding paths, sending the original image $x$ into both a low-resolution global view and high-resolution sliding window regional views.

(1) \textbf{Global View.} The image $x$ is resized into a global view $x^{(0)}$, which is then encoded to obtain an image-level token capturing coarse-grained semantics. We denote the process as:
$$s^{(0)} = f(x^{(0)}).$$

(2) \textbf{Sliding-window Region View.} At the original resolution, the image $x$ is partitioned into $L$ high-resolution patches $\{x^{(l)}\}_{l=1}^L$ via a sliding window with fixed size and stride. Each patch is independently encoded into region-level tokens capturing fine-grained semantics. We denote the process as:
$$s^{(l)} = f(x^{(l)}), \quad l = 1, \ldots, L.$$

The above $1 + L$ tokens are organized into a hierarchical token sequence in a coarse-to-fine order, denoted as:
$$s = \{s^{(0)}, s^{(1)}, \ldots, s^{(L)}\}.$$

The overall encoding process is denoted as:
$$s = f^{\mathrm{HSE}}(x).$$

Each client obtains its local set of real token sequences via HSE, denoted as $\mathcal{S}_n^{\mathrm{real}}$.

\subsection{M-AR Transformer Generator}

\paragraph{Conditional Modeling with Multi-scale Autoregression.}
To reflect the semantic dependencies from global to local and to reduce the complexity of modeling high-dimensional joint distributions, FALCON factorizes the class-conditional token sequence distribution using a multi-scale autoregressive (M-AR) approach:
\begin{equation}
p(s \mid y)
= p\bigl(s^{(0)} \mid y\bigr)\,\cdot\,p\bigl(s^{(1)},\dots,s^{(L)} \mid y,\,s^{(0)}\bigr).
\label{eq:m-ar}
\end{equation}

\begin{figure}[t]
\centering
\includegraphics[width=0.98\columnwidth]{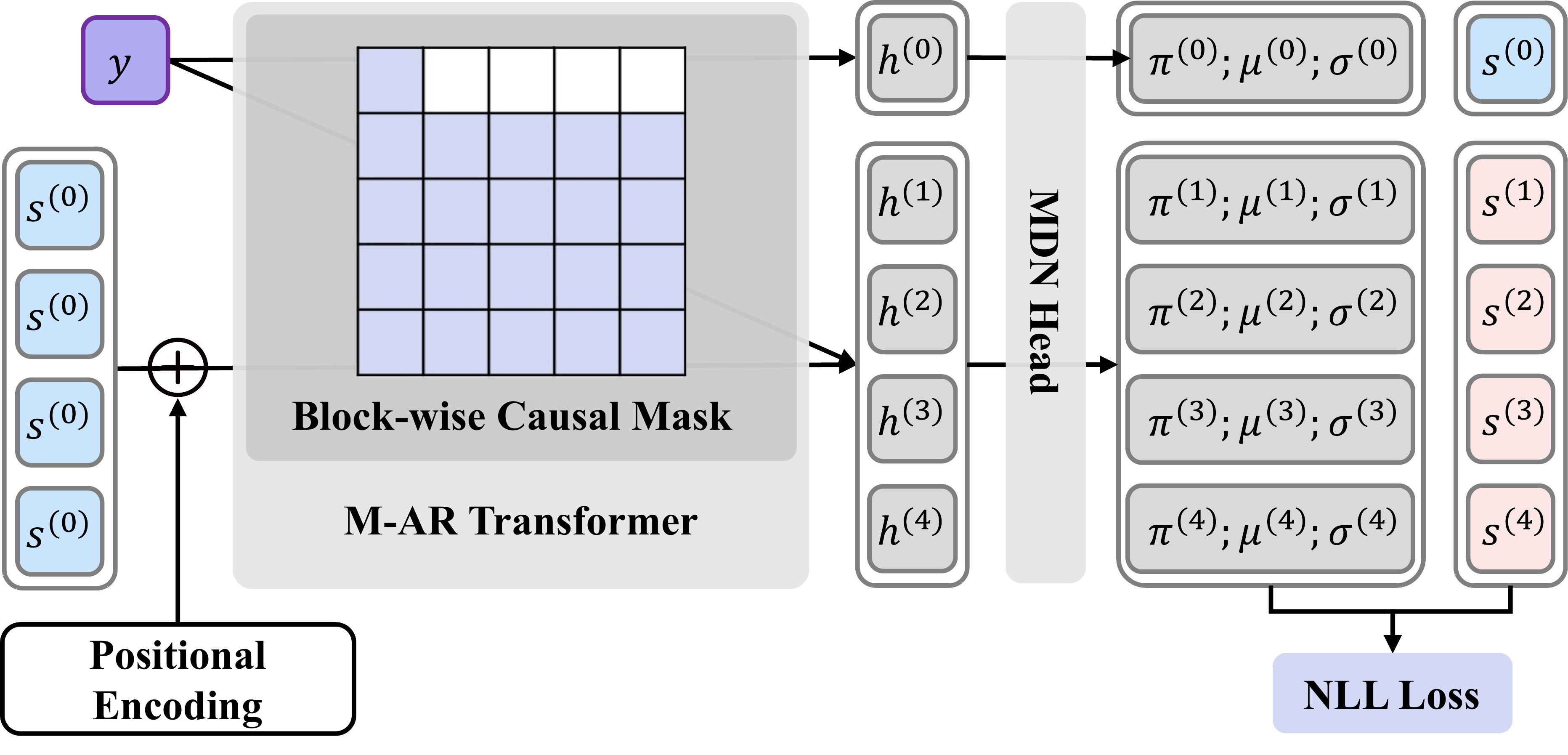}
\caption{Architecture of the M-AR Transformer Generator.}
\label{fig:M-AR}
\end{figure}

\paragraph{Conditional GMM Formulation.}
To instantiate the above hierarchical conditional factorization, we formulate the conditional distributions of image-level and region-level tokens using Gaussian Mixture Models (GMMs). Specifically, given a class label $y$, the conditional distribution of the image-level token $s^{(0)}$ is assumed to follow a GMM with $K$ components:
\begin{equation}
p\bigl(s^{(0)} \mid y\bigr)
= \sum_{k=1}^{K} \pi_{k}^{(0)} \,\mathcal{N}\bigl(s^{(0)} \mid \boldsymbol{\mu}_{k}^{(0)},\,\boldsymbol{\Sigma}_{k}^{(0)}\bigr),
\label{eq:image-level-gmm}
\end{equation}
where $\{\pi_{k}^{(0)},\,\boldsymbol{\mu}_{k}^{(0)},\,\boldsymbol{\Sigma}_{k}^{(0)}\}_{k=1}^{K}$ denote the mixture weights, means, and covariances of the GMM, respectively. 

Subsequently, conditioned on $s^{(0)}$ and $y$, we model the distribution of the $L$ region-level tokens $\left(s^{(1)},\dots,s^{(L)}\right)$ independently for each token position as:
\begin{multline}
p(s^{(1)}, \dots, s^{(L)} \mid s^{(0)}, y) = \\
\prod_{l=1}^L \sum_{k=1}^K \pi_k^{(l)} \,
\mathcal{N}\bigl(s^{(l)} \mid \boldsymbol{\mu}_k^{(l)}, \boldsymbol{\Sigma}_k^{(l)}\bigr),
\label{eq:reg_tokens}
\end{multline}
where, for each token position $l$, $\left\{\pi_k^{(l)},\,\boldsymbol{\mu}_k^{(l)},\,\boldsymbol{\Sigma}_k^{(l)}\right\}_{k=1}^{K}$ denote the mixture weights, means, and covariances for the $k$-th component of the GMM.

\paragraph{M-AR Transformer Architecture.}
As shown in Figure~\ref{fig:M-AR}, we design a M-AR Transformer generator $g(\cdot,; \theta^g)$, with each client $n$ maintaining a local generator $g_n(\cdot,; \theta^g_n)$.
The core architecture consists of a Transformer equipped with a block-wise causal mask~\cite{var24}, which enforces a coarse-to-fine attention mechanism across the token sequence.
Combining the multi-scale autoregressive factorization (see Eq.\ref{eq:m-ar}) with conditional GMM modeling for both image-level and region-level tokens (see Eqs.\ref{eq:image-level-gmm} and~\ref{eq:reg_tokens}), the Transformer models the sequential dependencies among tokens at different semantic scales.
At each position $t$ in the sequence, the hidden representation $h^{(t)}$ is passed to a Mixture Density Network (MDN) head, which outputs the GMM parameters for that position.

\paragraph{Training.}
During training, each client $n$ constructs an input sequence of length $1+L$ for each sample: $\{y, s^{(0)}, \ldots, s^{(L)}\}$. Self-attention is computed using a block-wise causal mask, and the MDN head predicts the GMM parameters at each position. This results in a locally estimated distribution $\hat{p}_n$. The training objective is to minimize the negative log-likelihood (NLL) loss:
\begin{align}
\mathcal{L}_{\mathrm{NLL}}^{(n)}(\theta_n^g) 
&= -\log \hat{p}_n \bigl(s^{(0)} \mid y;\, \theta_n^g \bigr) \nonumber\\
&\quad\;\;\;\cdot \hat{p}_n \bigl(s^{(1)}, \ldots, s^{(L)} \mid y,\, s^{(0)};\, \theta_n^g \bigr).
\label{eq:nll-loss}
\end{align}

\paragraph{Construction of Synthetic Token Sequence Set.}
Synthetic token sequences $s^{\mathrm{syn}}_n$ are sampled from the estimated distribution $\hat{p}_n(\cdot;\theta_n^g)$.

\subsection{Local Classifier Training}
Each client trains a local classifier $h_n$ on the real token sequence set $\mathcal{S}_n^{\mathrm{real}}$ using a cross-entropy loss.

\subsection{Distillation-Guided Global Classifier Training}
The server receives all uploaded local classifiers $\{h_n\}_{n=1}^N$ and synthetic token sequence sets $\{\mathcal{S}_n^{\mathrm{syn}}\}_{n=1}^N$, and constructs a combined training set $\mathcal{S}^{\mathrm{syn}}$. It then trains the global classifier $h$ under the guidance of knowledge distillation.

\paragraph{Classifier Design.}
The classifier includes attention pooling and a classification head.Each input token sequence $s$ is first pooled using a learnable query vector $q \in \mathbb{R}^d$ to integrate multi-scale semantic information:
\begin{equation}
K = s W_k \in \mathbb{R}^{(1+L)\times d}, \quad
V = s W_v \in \mathbb{R}^{(1+L)\times d}.
\label{eq:key-value}
\end{equation}

The attention scores are computed as:
\begin{equation}
\alpha = \mathrm{softmax}\!\bigl(\tfrac{q K^T}{\sqrt{d}}\bigr).
\label{eq:attention}
\end{equation}

The semantic summary is obtained as:
\begin{equation}
s^{\mathrm{sem}} = \alpha V.
\label{eq:semantic-summary}
\end{equation}

Finally, the semantic summary $s^{\mathrm{sem}}$ is passed into the classification head $h_{\mathrm{cls}}(\cdot)$ to perform multi-class prediction:
\begin{equation}
\hat{y} = h_{\mathrm{cls}}(s^{\mathrm{sem}}).
\label{eq:classifier-pred}
\end{equation}

\paragraph{Knowledge Distillation Loss.}
Since the synthetic data only indirectly optimize the FL objective, we introduce a knowledge distillation mechanism that integrates the local classifiers as teachers to correct the optimization target and compensate for limitations in distribution modeling by the generator. Specifically, for each input token sequence $s$, we apply temperature scaling to the softmax outputs of all local classifiers $\{h_n\}_{n=1}^{N}$ on $s$, and compute their average to obtain the ensemble teacher distribution, defined as:
\begin{equation}
p_T(y \mid s) = \frac{1}{N} \sum_{n=1}^N \mathrm{softmax}\!\bigl(\tfrac{z_n(s)}{T}\bigr),
\label{eq:teacher-dist}
\end{equation}
where $z_n(s)$ denotes the logits output of the $n$-th local classifier on $s$, and $T$ is the distillation temperature.

The probability distribution output by the global model is:
\begin{equation}
p_S(y \mid s) = \mathrm{softmax}\!\left(\frac{z_S(s)}{T}\right),
\label{eq:student-dist}
\end{equation}
where $z_S(s)$ denotes the logits of the global classifier.

The distillation loss is defined as the Kullback–Leibler (KL) divergence between the ensemble teacher distribution and the global classifier distribution:
\begin{equation}
\ell_{\mathrm{KD}} = T^2 \cdot \mathrm{KL}\bigl(p_T(y \mid s) \,\|\, p_S(y \mid s)\bigr).
\label{eq:kd-loss}
\end{equation}

\paragraph{Overall Loss.}
The final loss function is a weighted combination of the cross-entropy loss and the distillation loss:
\begin{equation}
\mathcal{L}_{\mathrm{total}} = (1 - \alpha) \, \ell_{\mathrm{CE}} + \alpha \, \ell_{\mathrm{KD}},
\label{eq:total-loss}
\end{equation}
where $\ell_{\mathrm{CE}}$ is the standard cross-entropy loss and $\alpha$ is the balancing coefficient. Guided by distillation, the global classifier is able to integrate the knowledge captured by local classifiers on real token sequences from each client.

After training, the global classifier $h$ is distributed to all clients, forming the global model $w = f^{\mathrm{HSE}} \cdot h$.

\begin{table}[t]
\centering
\footnotesize
\setlength{\tabcolsep}{3.2pt}
\renewcommand{\arraystretch}{1.05}

\begin{tabular}{lcccc}
\toprule
\textbf{Dataset} & \textbf{Domain} & \textbf{Non-IID Type} & \textbf{\#Classes} & \textbf{\#Clients} \\
\midrule
Tuberculosis    & Med. & Feat. & 2   & 3 \\
PACS            & Nat. & Feat. & 7   & 4 \\
OfficeHome      & Nat. & Feat. & 65  & 4 \\
\midrule
PMRAM           & Med. & Label & 4   & 5 \\
Pneumonia       & Med. & Label & 2   & 5 \\
Tiny-ImageNet   & Nat. & Label & 200 & 5 \\
\bottomrule
\end{tabular}

\caption{Dataset summary. ``Med.''/``Nat.'' denote domains; ``Feat.''/``Label'' indicate types of non-IID.}
\label{tab:dataset}
\end{table}

\begin{table*}[t]
\centering
\resizebox{\textwidth}{!}{%
\begin{tabular}{lccc ccc ccc ccc c}  
\toprule
\multirow{3}{*}{}
    & \multicolumn{3}{c}{Feature non-IID}
    & \multicolumn{9}{c}{Label non-IID}
    & \multirow{3}{*}{Avg.} \\
\cmidrule(lr){2-4} \cmidrule(lr){5-13}
    & Tuberculosis & PACS & OfficeHome
    & \multicolumn{3}{c}{PMRAM}
    & \multicolumn{3}{c}{Pneumonia}
    & \multicolumn{3}{c}{Tiny-ImageNet}
    & \\
\cmidrule(lr){5-7} \cmidrule(lr){8-10} \cmidrule(lr){11-13}
    &   &   &
    & 0.1 & 0.3 & 0.5
    & 0.1 & 0.3 & 0.5
    & 0.1 & 0.3 & 0.5
    & \\
\midrule
FedAvg  & 85.07 & 87.61 & 82.51 & 90.98 & 94.20 & 95.83 & 66.11 & 86.54 & 88.83 & 30.75 & 34.70 & 36.51 & 73.30 \\
\midrule
DENSE   & \underline{76.48} & 51.63 & 60.74 & 26.93 & 26.53 & 54.15 & 64.12 & 78.75 & \underline{87.17} & 21.70 & 26.87 & 30.90 & 50.50 \\
FedLMG  & 47.32 & 61.42 & 60.19 & 36.99 & 36.54 & 37.98 & 57.87 & 59.97 & 63.09 & 26.83 & 30.22 & 31.41 & 45.82 \\
FedPFT  & 72.62 & 96.03 & 83.85 & \underline{74.09} & \underline{74.38} & \underline{77.74} & 72.65 & 75.27 & 76.74 & 59.40 & 59.70 & 60.78 & 73.61 \\
FedCGS  & 53.50 & \textbf{97.50} & \textbf{90.16} & 64.72 & 69.04 & 62.79 & \underline{77.62} & \underline{79.12} & 76.90 & \textbf{81.53} & \underline{81.53} & \underline{81.53} & \underline{76.34} \\
\midrule
FALCON & \textbf{87.20} & \underline{96.42} & \underline{89.68} & \textbf{83.33} & \textbf{85.60} & \textbf{85.16} & \textbf{82.41} & \textbf{88.92} & \textbf{89.22} & \underline{79.93} & \textbf{81.59} & \textbf{81.55} & \textbf{85.92} \\
\bottomrule
\end{tabular}%
}
\caption{Comparison of test accuracy (\%) under feature and label non-IID settings. 
The first three columns show results on feature non-IID datasets; 
the next nine columns report results under label non-IID scenarios with Dirichlet distribution ($\alpha \in \{0.1, 0.3, 0.5\}$). 
The final column ("Avg.") denotes the average across all settings. 
Among OSFL methods, the best and second-best results are highlighted in \textbf{bold} and \underline{underline}, respectively.}
\label{tab:results}
\end{table*}

\begin{table}[t]
\centering
\resizebox{\columnwidth}{!}{%
\begin{tabular}{lccc|cc}
\toprule
Dataset & Full(M-AR) & w/o HSE & w/o KD & AR & NAR \\
\midrule
Tuberculosis         & 87.20 & 83.63 & 84.23 & 86.61 & 86.90 \\
PACS                 & 96.42 & 96.16 & 95.71 & 96.29 & 96.23 \\
OfficeHome           & 89.68 & 89.05 & 86.87 & 84.26 & 86.98 \\
\midrule
PMRAM                & 84.70 & 81.26 & 80.41 & 83.38 & 78.67 \\
Pneumonia & 86.85 & 82.12 & 84.42 & 72.76 & 85.73 \\
Tiny-ImageNet        & 81.02 & 79.59 & 77.20 & 80.70 & 80.53 \\
\midrule
Avg.                 & 85.92 & 82.32 & 82.82 & 81.47 & 82.99 \\
\bottomrule
\end{tabular}
}
\caption{
Ablation study of FALCON. 
``Full'' denotes the complete method; ``w/o HSE'' disables hierarchical scale encoding; ``w/o KD'' disables knowledge distillation. 
``AR'' and ``NAR'' represent the vanilla autoregressive and non-autoregressive generator variants, respectively. 
}
\vspace{-0.1in}
\label{tab:ablation-transpose}
\end{table}

\section{Experiments and Analyses}
\subsection{Experimental Settings}
\paragraph{Datasets.}

Table~\ref{tab:dataset} summarizes the datasets used for evaluating federated learning performance under both feature and label non-IID scenarios, covering natural and medical imaging domains.
For the feature non-IID setting, we select datasets with inherent multi-source structures and treat each original subset as a separate client in the federated setup. Tuberculosis~\cite{mpcpa24} consists of chest X-ray images collected from three hospitals in Montgomery\cite{mc}, China\cite{sz}, and India. PACS~\cite{pacs} and OfficeHome~\cite{officehome} each contain images from four visual domains: Art Painting, Cartoon, Photo, and Sketch for PACS; and Art, Clipart, Product, and Real-World for OfficeHome.
For label non-IID settings, we simulate heterogeneous label distributions across clients using PMRAM~\cite{pmram}, Pneumonia~\cite{xray}, and Tiny-ImageNet~\cite{tiny}. In these cases, we construct five clients by partitioning each dataset with a Dirichlet distribution over class labels to control label heterogeneity. 
To accommodate differences in dataset scale, we employ a 6:2:2 split for training, validation, and test sets in PMRAM, whereas Pneumonia adopts an 8:2 split for training and validation with the official test set for evaluation. All other datasets use a 8:1:1 split for training, validation, and testing. All images are resized to $448 \times 448$ resolution.

\paragraph{Baselines.}
To evaluate the effectiveness of the proposed method, we compare FALCON with FedAvg \cite{avg17} as well as several representative approaches from the current one-shot federated learning (OSFL) literature, including DENSE \cite{dense22}, FedLMG \cite{lmg25}, FedPFT \cite{pft}, and FedCGS \cite{cgs25}. Specifically, DENSE serves as a classical data-free knowledge distillation baseline. FedLMG employs a pretrained diffusion model with classifier guidance to generate synthetic data. Both FedPFT and FedCGS are methods based on semantic feature distribution modeling.

\paragraph{Configurations.}
We use DINOv2-base \cite{dino} as the pretrained visual encoder. For hierarchical scale encoding (HSE), both the sliding window size and stride are set to 224. The token sequence generator adopts a 4-layer architecture with a hidden dimension of 768 and 16 attention heads, using diagonal covariance and $K=20$. It is trained for 400 epochs with a batch size of 256, using the AdamW optimizer with a learning rate of $1\text{e-}4$ and a weight decay of $1\text{e-}4$.
The local classifier is trained for 60 epochs with a batch size of 256 using the Adam optimizer and a learning rate of $5\text{e-}4$. For global classifier training, the balancing coefficient $\alpha$ is set to 0.5 and the distillation temperature $T$ is 4. The global classifier is trained for 60 epochs with a batch size of 256, using Adam and a learning rate of $5\text{e-}4$.
For the baselines, FedAvg, DENSE, and FedLMG use a local training batch size of 64, trained for 60 epochs with the Adam optimizer and a learning rate of $5\text{e-}4$. FedPFT uploads the parameters of a diagonal-covariance Gaussian mixture model $\mathcal{G}(K)$ with $K=20$, and all other settings follow the original paper. FedCGS also uses the DINOv2-base encoder, with all other settings consistent with the original implementation.

\subsection{Main Results}
Table~\ref{tab:results} reports the classification accuracy (\%) of each method under both feature and label non-IID scenarios. FALCON achieves the best or second-best results across all datasets and settings among OSFL methods. Specifically:
(1) \textbf{Strong Performance under Feature non-IID.} FALCON outperforms the best OSFL baseline by 10.72\% on the Tuberculosis dataset, with a performance gain of up to 39.88\% compared to the weakest baseline. It also maintains consistently strong results on PACS and OfficeHome. 
(2) \textbf{Stable Advantage under Label non-IID.} While the performance of existing OSFL methods tends to fluctuate or degrade as the Dirichlet parameter $\alpha$ decreases, FALCON maintains consistently high accuracy across all $\alpha$ values and datasets, demonstrating strong robustness to varying degrees of label heterogeneity.
(3) \textbf{Superior Results on Medical datasets.} Across six label non-IID settings on the PMRAM and Pneumonia datasets, FALCON consistently achieves the best results among OSFL methods. In these settings, FALCON outperforms the best OSFL baseline by an average of 7.83\%, with improvements of 7.42\%–11.22\% on PMRAM and 2.05\%–9.80\% on Pneumonia. These results demonstrate the advantages of our hierarchical scale encoding for modeling high-resolution regional details in medical images and effectively leveraging multi-scale information from raw images.
(4) \textbf{Reliable and Robust Overall Performance.} FALCON delivers stable and robust performance, achieving an average improvement of 9.58\% over the strongest OSFL baseline across all settings. In contrast to other methods, which tend to suffer severe performance degradation on certain datasets or non-IID configurations, FALCON maintains high accuracy in all cases. For example, on the Tuberculosis dataset, the second-best method, FedCGS, only reaches 53.50\% accuracy, whereas FALCON achieves 87.20\%. Similarly, for the PMRAM dataset, over half of the competing baselines (e.g., DENSE, FedLMG) fall below 40\% accuracy, while FALCON remains above 80\% across all $\alpha$ settings.

\subsection{Computational Efficiency of Sample Generation}

\begin{figure}[t]
\centering
\includegraphics[width=1\columnwidth]{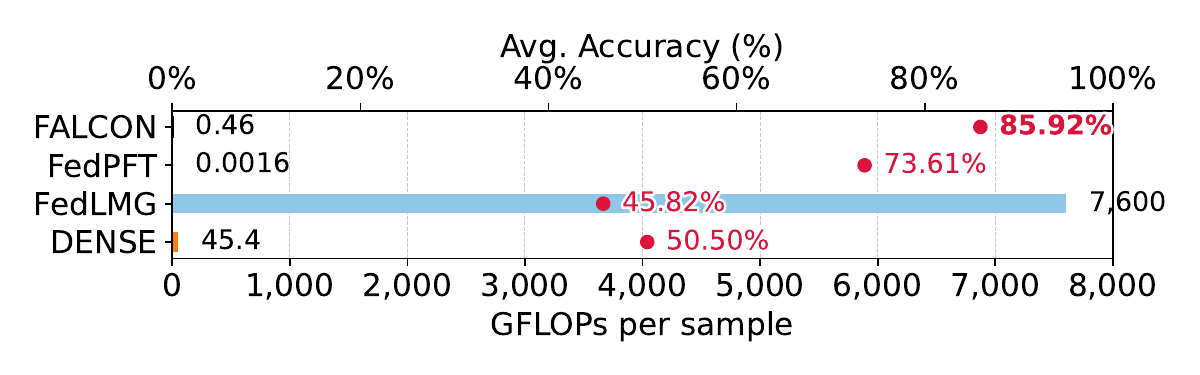}  
\caption{
Computational cost (bars, GFLOPs per sample) and average classification accuracy (dots, \%) for OSFL methods involving sample generation.
}
\label{fig:compute_cost}
\vspace{-0.1in}
\end{figure}

Figure~\ref{fig:compute_cost} compares the per-sample computational cost (in GFLOPs) and average classification accuracy (\%) of OSFL methods involving sample generation, including DENSE, which generates pseudo-samples; FedLMG, which leverages a pretrained diffusion model; FedPFT, which performs direct sampling from a GMM; and our proposed method, FALCON. Notably, FedLMG incurs an extremely high cost (7,600 GFLOPs per sample) due to the expensive multi-step sampling in diffusion-based generation. In contrast, FALCON reduces the per-sample cost to 0.46 GFLOPs, which is over four orders of magnitude lower, while ensuring strong expressive capacity. Although FedPFT achieves the lowest computational cost by directly sampling from a statistical feature distribution, it lacks the expressive capacity and generation fidelity of transformer-based sequence modeling. These results underscore FALCON’s computational efficiency, making it well-suited for resource-constrained or large-scale federated deployments.

\begin{table}[t]
\centering
\small
\begin{tabular}{lccc}
\toprule
        & Tuberculosis & PACS & OfficeHome \\
\midrule
M-AR        & 0.003227 & 0.002047 & 0.000905 \\
AR         & 0.003868 & 0.002457 & 0.001079 \\
NAR        & 0.003868 & 0.002151 & 0.000849 \\
\bottomrule
\end{tabular}
\caption{Maximum mean discrepancy (MMD) between real and synthetic token sequences on feature non-IID datasets, comparing different generator variants. Lower is better.}
\label{tab:mmd}
\vspace{-0.1in}
\end{table}

\subsection{Ablation Studies}
We conducted ablation studies to investigate the effects of key components, including hierarchical scale encoding (HSE), the token sequence generator architecture, and knowledge distillation. Due to space constraints, for label non-IID datasets, we report the mean accuracy across different Dirichlet parameters ($\alpha$) for each dataset. Table~\ref{tab:results} reports the results of the ablation study.

\paragraph{Impact of Hierarchical Scale Encoding (HSE).}
Introducing HSE consistently improves performance across all datasets, raising the overall average by 3.60\%. The improvement is particularly notable for the medical imaging benchmarks Tuberculosis, PMRAM, and Pneumonia, with improvements of 3.57\%, 3.44\%, and 4.73\%, respectively. This demonstrates the effectiveness of HSE in capturing fine-grained information essential for complex recognition tasks.

\paragraph{Impact of Token Sequence Generator Architecture.}
The M-AR generator we adopt in FALCON achieves the highest performance among all generator variants.
To further quantify the ability of each generator variant to approximate the token sequence distribution, we compute the maximum mean discrepancy (MMD) between real and synthetic token sequences on the feature non-IID datasets. We report results on Tuberculosis, PACS, and OfficeHome in Table~\ref{tab:mmd}. Our M-AR generator achieves the lowest MMD on most datasets, indicating a closer match to the real distribution compared to the AR and NAR generators.
These results demonstrate that modeling dependencies among hierarchical tokens is critical for capturing the token sequence distribution.

\paragraph{Impact of Knowledge Distillation.}
Incorporating knowledge distillation yields a substantial improvement in average accuracy, increasing from 82.82\% to 85.92\%, with the largest observed improvement on a single dataset reaching 5.52\%. These results highlight the critical role of knowledge distillation in enhancing global model performance.

\subsection{Evaluating under Reconstruction Attacks}

\begin{figure}[t]
\centering
\includegraphics[width=1\columnwidth]{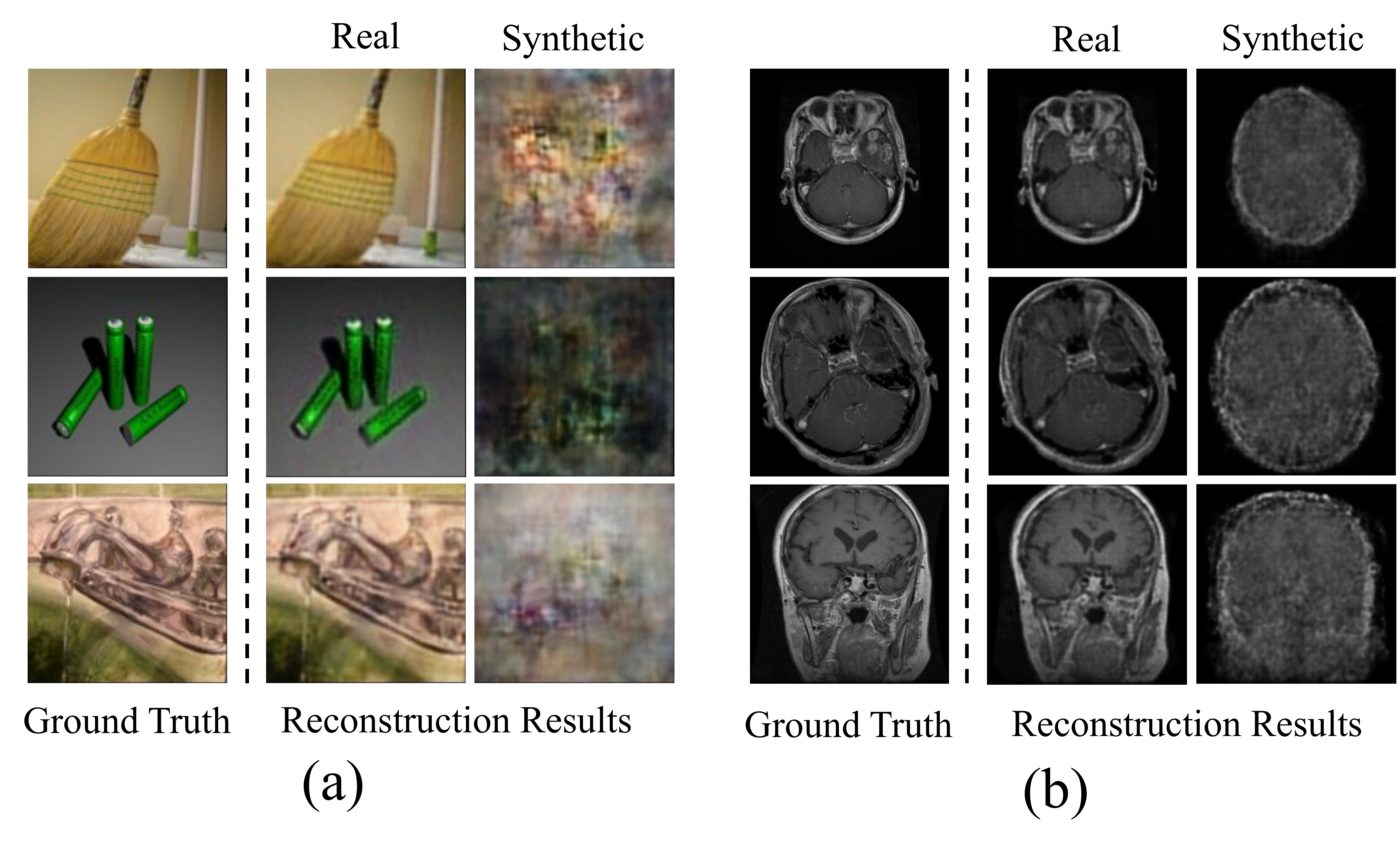}  
\vspace{-0.3in}
\caption{
        Visualization of reconstruction attack. 
        (a) OfficeHome ; (b) PMRAM ($\alpha=0.5$).
        For each, left: original images; middle: reconstructions from real token sequences; right: reconstructions from synthetic token sequences.
        }
\label{fig:recon}
\vspace{-0.1in}
\end{figure}

To evaluate potential privacy leakage, we design a reconstruction attack by training a conditional generative model to recover original images from token sequences, applying the attack to real and synthetic sequences of a selected client.

Figure~\ref{fig:recon} visualizes reconstruction results for both OfficeHome and PMRAM ($\alpha=0.5$). For synthetic token sequences, we present the reconstruction with the highest SSIM among all generated samples for each original image, corresponding to a conservative, worst-case evaluation of potential privacy risk. 
When using real token sequences, the reconstructed images retain substantial correspondence to the original inputs, indicating that detailed visual information can still be recovered. In contrast, reconstructions from synthetic token sequences are largely unrecognizable, exhibiting significant loss of structural and semantic content, demonstrating strong privacy protection.

For quantitative evaluation, we report the average peak signal-to-noise ratio (PSNR) and structural similarity index (SSIM) for each dataset. For real token sequences, the average similarity between original and reconstructed images is computed directly. For synthetic token sequences, due to the lack of one-to-one correspondence, we report the highest PSNR and SSIM across all reconstructions for each image, and average these values. As shown in Table~\ref{tab:recon-quality}, constructions from synthetic tokens yield much lower similarity scores on both datasets, indicating the privacy-preserving advantage of FALCON. Notably, the SSIM for synthetic tokens is higher on PMRAM than on OfficeHome, consistent with the greater homogeneity and structural repetition of medical images. This facilitates generative modeling of global anatomical features, yet does not compromise privacy, as patient-specific details remain unrecognizable.

\begin{table}[t]
\centering
\footnotesize
\begin{tabular}{lcc|cc}
\toprule
\multirow{2}{*}{\textbf{Token Source}} & \multicolumn{2}{c|}{\textbf{OfficeHome}} & \multicolumn{2}{c}{\textbf{PMRAM} ($\alpha=0.5$)} \\
 & PSNR & SSIM & PSNR & SSIM \\
\midrule
Real      & 27.18  & 0.8111 & 28.75 & 0.8176 \\
Synthetic & 13.30  & 0.1534 & 18.37 & 0.4368 \\
\bottomrule
\end{tabular}
\caption{
Average PSNR and SSIM for reconstructions from real and synthetic token sequences on OfficeHome and PMRAM ($\alpha=0.5$).
}
\label{tab:recon-quality}
\vspace{-0.1in}
\end{table}

\subsection{Communication Cost Analysis}
In FALCON, each client can either upload the synthetic token sequences or the generator model, depending on which option yields a lower communication cost under the given configuration. In our experiments, the generator model size is approximately 144.33 MB, and each $5 \times 768$ token sequence occupies about 15 KB with 32-bit storage. Uploading token sequences is thus more efficient when the total number of sequences is less than about 9,850. The per-client communication cost is therefore bounded by the combined size of the generator and local classifier, reaching up to 166.86 MB in the worst case. 
In practice, the actual cost is often substantially lower, as FALCON enables clients to flexibly select the upload strategy that best suits their data volume and communication constraints.


\section{Conclusion}
We propose FALCON, a one-shot federated learning framework tailored for diverse application domains and multi-type non-IID scenarios. FALCON introduces HSE to extract hierarchical token sequences without fine-tuning the encoder, and employs an M-AR transformer to model the distribution of token sequences. Combined with distillation-guided global training, the server learns a high-performing classifier. 
Extensive experiments show that FALCON delivers robust and competitive results across diverse tasks, consistently matching or surpassing the best existing OSFL approaches.
Our method significantly enhances the practical utility of pretrained encoders for federated learning in real-world settings, improving semantic retention and demonstrating strong scalability. These advantages position FALCON as a promising paradigm for federated classification.
To further reinforce these benefits, key directions for future work are to improve encoding effectiveness while ensuring all clients adopt a consistent encoding strategy, and to explore more powerful generative modeling paradigms with improved effectiveness and efficiency.

\section*{Acknowledgments}
This work is supported by National Key Research and Development Program of China (2024YFE0203100), Guangdong Provincial Key Laboratory of Ultra High Definition Immersive Media Technology (Grant No. 2024B1212010006), Shenzhen Science and Technology Program (JCYJ20230807120800001), and the project supplementary funding of National Innovation 2030 Major S\&T Project of China (2020AAA0104203).

\bibliography{aaai2026}

@article{lyu2020threats,
  title={Threats to federated learning: A survey},
  author={Lyu, Lingjuan and Yu, Han and Yang, Qiang},
  journal={arXiv preprint arXiv:2003.02133},
  year={2020}
}

@inproceedings{nasr2019comprehensive,
  title={Comprehensive privacy analysis of deep learning: Passive and active white-box inference attacks against centralized and federated learning},
  author={Nasr, Milad and Shokri, Reza and Houmansadr, Amir},
  booktitle={IEEE symposium on security and privacy},
  pages={739--753},
  year={2019},
}

@ARTICLE{24poison,
  author={Yazdinejad, Abbas and Dehghantanha, Ali and Karimipour, Hadis and Srivastava, Gautam and Parizi, Reza M.},
  journal={IEEE Transactions on Information Forensics and Security}, 
  title={A Robust Privacy-Preserving Federated Learning Model Against Model Poisoning Attacks}, 
  year={2024},
  volume={19},
  number={},
  pages={6693-6708},
  doi={10.1109/TIFS.2024.3420126}}

@ARTICLE{25poison,
  author={Rao, Bosen and Zhang, Jiale and Wu, Di and Zhu, Chengcheng and Sun, Xiaobing and Chen, Bing},
  journal={IEEE Transactions on Artificial Intelligence}, 
  title={Privacy Inference Attack and Defense in Centralized and Federated Learning: A Comprehensive Survey}, 
  year={2025},
  volume={6},
  number={2},
  pages={333-353},
  doi={10.1109/TAI.2024.3363670}}

@article{guha2019,
  author       = {Neel Guha and
                  Ameet Talwalkar and
                  Virginia Smith},
  title        = {One-Shot Federated Learning},
  journal      = {CoRR},
  volume       = {abs/1902.11175},
  year         = {2019},
  url          = {http://arxiv.org/abs/1902.11175},
  eprinttype    = {arXiv},
  eprint       = {1902.11175},
  bibsource    = {dblp computer science bibliography, https://dblp.org}
}

@inproceedings{fusefl24,
  author = {Tang, Zhenheng and Zhang, Yonggang and Dong, Peijie and Cheung, Yiu-ming and Zhou, Amelie Chi and Han, Bo and Chu, Xiaowen},
  title        = {FuseFL: One-Shot Federated Learning through the Lens of Causality with Progressive Model Fusion},
  booktitle    = {Advances in Neural Information Processing Systems},
  year         = {2024},
  volume       = {37},
  pages = {28393--28429},
  url          = {http://papers.nips.cc/paper\_files/paper/2024/hash/31e6e0c09325a3be16d93f84e40e0c7e-Abstract-Conference.html},
}

@InProceedings{fisher24,
  title = 	 {{FedFisher}: Leveraging {F}isher Information for One-Shot Federated Learning},
  author =       {Jhunjhunwala, Divyansh and Wang, Shiqiang and Joshi, Gauri},
  booktitle = 	 {Proceedings of The 27th International Conference on Artificial Intelligence and Statistics},
  pages = 	 {1612--1620},
  year = 	 {2024},
  editor = 	 {Dasgupta, Sanjoy and Mandt, Stephan and Li, Yingzhen},
  volume = 	 {238},
  series = 	 {Proceedings of Machine Learning Research},
  month = 	 {02--04 May},
  publisher =    {PMLR},
  url = 	 {https://proceedings.mlr.press/v238/jhunjhunwala24a.html},
}

@inproceedings{fedbe21,
  title={FedBE: Making Bayesian Model Ensemble Applicable to Federated Learning},
  author={Chen, Hong-You and Chao, Wei-Lun},
  booktitle = {ICLR},
  year={2021}
}

@article{bpb24, title={Calibrated One Round Federated Learning with Bayesian Inference in the Predictive Space}, volume={38}, url={https://ojs.aaai.org/index.php/AAAI/article/view/29122}, DOI={10.1609/aaai.v38i11.29122}, number={11}, journal={Proceedings of the AAAI Conference on Artificial Intelligence}, author={Hasan, Mohsin and Zhang, Guojun and Guo, Kaiyang and Chen, Xi and Poupart, Pascal}, year={2024}, month={Mar.}, pages={12313-12321} }

@inproceedings{fens24,
 author = {Allouah, Youssef and Dhasade, Akash and Guerraoui, Rachid and Gupta, Nirupam and Kermarrec, Anne-Marie and Pinot, Rafael and Pires, Rafael and Sharma, Rishi},
 booktitle = {Advances in Neural Information Processing Systems},
 editor = {A. Globerson and L. Mackey and D. Belgrave and A. Fan and U. Paquet and J. Tomczak and C. Zhang},
 pages = {68500--68527},
 publisher = {Curran Associates, Inc.},
 title = {Revisiting Ensembling in One-Shot Federated Learning},
 url = {https://proceedings.neurips.cc/paper_files/paper/2024/file/7ea46207ec9bda974b140fe11d8dd727-Paper-Conference.pdf},
 volume = {37},
 year = {2024}
}

@inproceedings{feddf20,
 author = {Lin, Tao and Kong, Lingjing and Stich, Sebastian U and Jaggi, Martin},
 booktitle = {Advances in Neural Information Processing Systems},
 editor = {H. Larochelle and M. Ranzato and R. Hadsell and M.F. Balcan and H. Lin},
 pages = {2351--2363},
 publisher = {Curran Associates, Inc.},
 title = {Ensemble Distillation for Robust Model Fusion in Federated Learning},
 url = {https://proceedings.neurips.cc/paper_files/paper/2020/file/18df51b97ccd68128e994804f3eccc87-Paper.pdf},
 volume = {33},
 year = {2020}
}

@inproceedings{dense22,
 author = {Zhang, Jie and Chen, Chen and Li, Bo and Lyu, Lingjuan and Wu, Shuang and Ding, Shouhong and Shen, Chunhua and Wu, Chao},
 booktitle = {Advances in Neural Information Processing Systems},
 editor = {S. Koyejo and S. Mohamed and A. Agarwal and D. Belgrave and K. Cho and A. Oh},
 pages = {21414--21428},
 publisher = {Curran Associates, Inc.},
 title = {DENSE: Data-Free One-Shot Federated Learning},
 url = {https://proceedings.neurips.cc/paper_files/paper/2022/file/868f2266086530b2c71006ea1908b14a-Paper-Conference.pdf},
 volume = {35},
 year = {2022}
}

@InProceedings{isca23,
author="Kang, Myeongkyun
and Chikontwe, Philip
and Kim, Soopil
and Jin, Kyong Hwan
and Adeli, Ehsan
and Pohl, Kilian M.
and Park, Sang Hyun",
editor="Greenspan, Hayit
and Madabhushi, Anant
and Mousavi, Parvin
and Salcudean, Septimiu
and Duncan, James
and Syeda-Mahmood, Tanveer
and Taylor, Russell",
title="One-Shot Federated Learning on Medical Data Using Knowledge Distillation with Image Synthesis and Client Model Adaptation",
booktitle="Medical Image Computing and Computer Assisted Intervention -- MICCAI 2023",
year="2023",
publisher="Springer Nature Switzerland",
address="Cham",
pages="521--531",
}

@inproceedings{coboosting24,
title={Enhancing One-Shot Federated Learning Through Data and Ensemble Co-Boosting},
author={Rong Dai and Yonggang Zhang and Ang Li and Tongliang Liu and Xun Yang and Bo Han},
booktitle={The Twelfth International Conference on Learning Representations},
year={2024},
url={https://openreview.net/forum?id=tm8s3696Ox}
}

@article{dosfl20,
  title={Distilled one-shot federated learning},
  author={Zhou, Yanlin and Pu, George and Ma, Xiyao and Li, Xiaolin and Wu, Dapeng},
  journal={arXiv preprint arXiv:2009.07999},
  year={2020}
}

@inproceedings{fedd323,
  title={Federated learning via decentralized dataset distillation in resource-constrained edge environments},
  author={Song, Rui and Liu, Dai and Chen, Dave Zhenyu and Festag, Andreas and Trinitis, Carsten and Schulz, Martin and Knoll, Alois},
  booktitle={2023 International Joint Conference on Neural Networks (IJCNN)},
  pages={1--10},
  year={2023},
  organization={IEEE}
}

@article{sd2c24,
  title={One-shot federated learning via synthetic distiller-distillate communication},
  author={Zhang, Junyuan and Liu, Songhua and Wang, Xinchao},
  journal={Advances in Neural Information Processing Systems},
  volume={37},
  pages={102611--102633},
  year={2024}
}

@article{osgan23,
  title={OSGAN: One-shot distributed learning using generative adversarial networks: A. Kasturi, C. Hota},
  author={Kasturi, Anirudh and Hota, Chittaranjan},
  journal={The Journal of Supercomputing},
  volume={79},
  number={12},
  pages={13620--13640},
  year={2023},
  publisher={Springer}
}

@inproceedings{fedcvae23,
  title={Data-free one-shot federated learning under very high statistical heterogeneity},
  author={Heinbaugh, Clare Elizabeth and Luz-Ricca, Emilio and Shao, Huajie},
  booktitle={The Eleventh International Conference on Learning Representations},
  year={2023}
}

@misc{fgl23,
      title={Federated Generative Learning with Foundation Models}, 
      author={Jie Zhang and Xiaohua Qi and Bo Zhao},
      year={2023},
      eprint={2306.16064},
      archivePrefix={arXiv},
      primaryClass={cs.LG}
}

@inproceedings{feddeo24,
  title={Feddeo: Description-enhanced one-shot federated learning with diffusion models},
  author={Yang, Mingzhao and Su, Shangchao and Li, Bin and Xue, Xiangyang},
  booktitle={Proceedings of the 32nd ACM International Conference on Multimedia},
  pages={6666--6675},
  year={2024}
}

@inproceedings{fedbip25,
  title={Fedbip: Heterogeneous one-shot federated learning with personalized latent diffusion models},
  author={Chen, Haokun and Li, Hang and Zhang, Yao and Bi, Jinhe and Zhang, Gengyuan and Zhang, Yueqi and Torr, Philip and Gu, Jindong and Krompass, Denis and Tresp, Volker},
  booktitle={Proceedings of the Computer Vision and Pattern Recognition Conference},
  pages={30440--30450},
  year={2025}
}

@inproceedings{lmg25,
  title={One-Shot Heterogeneous Federated Learning with Local Model-Guided Diffusion Models},
  author={Yang, Mingzhao and Su, Shangchao and Li, Bin and Xue, Xiangyang},
  booktitle={Forty-second International Conference on Machine Learning},
  year={2025}
}

@article{gan20,
  title={Generative adversarial networks},
  author={Goodfellow, Ian and Pouget-Abadie, Jean and Mirza, Mehdi and Xu, Bing and Warde-Farley, David and Ozair, Sherjil and Courville, Aaron and Bengio, Yoshua},
  journal={Communications of the ACM},
  volume={63},
  number={11},
  pages={139--144},
  year={2020},
  publisher={ACM New York, NY, USA}
}

@misc{vae13,
  title={Auto-encoding variational bayes},
  author={Kingma, Diederik P and Welling, Max and others},
  year={2013},
  publisher={Banff, Canada}
}

@article{ddpm20,
  title={Denoising diffusion probabilistic models},
  author={Ho, Jonathan and Jain, Ajay and Abbeel, Pieter},
  journal={Advances in neural information processing systems},
  volume={33},
  pages={6840--6851},
  year={2020}
}

@inproceedings{ldm22,
  title={High-resolution image synthesis with latent diffusion models},
  author={Rombach, Robin and Blattmann, Andreas and Lorenz, Dominik and Esser, Patrick and Ommer, Bj{\"o}rn},
  booktitle={Proceedings of the IEEE/CVF conference on computer vision and pattern recognition},
  pages={10684--10695},
  year={2022}
}

@misc{wgan17,
      title={Wasserstein GAN}, 
      author={Martin Arjovsky and Soumith Chintala and Léon Bottou},
      year={2017},
      eprint={1701.07875},
      archivePrefix={arXiv},
      primaryClass={stat.ML},
      url={https://arxiv.org/abs/1701.07875}, 
}

@article{pft,
  title={Foundation Models Meet Federated Learning: A One-shot Feature-sharing Method with Privacy and Performance Guarantees},
  author={Beitollahi, Mahdi and Bie, Alex and Hemati, Sobhan and Brunswic, Leo Maxime and Li, Xu and Chen, Xi and Zhang, Guojun},
  journal={Transactions on Machine Learning Research},
  year={2025}
}

@inproceedings{cgs25,
  title={Capture global feature statistics for one-shot federated learning},
  author={Guan, Zenghao and Zhou, Yucan and Gu, Xiaoyan},
  booktitle={Proceedings of the AAAI Conference on Artificial Intelligence},
  volume={39},
  number={16},
  pages={16942--16950},
  year={2025}
}

@article{attn17,
  title={Attention is all you need},
  author={Vaswani, Ashish and Shazeer, Noam and Parmar, Niki and Uszkoreit, Jakob and Jones, Llion and Gomez, Aidan N and Kaiser, {\L}ukasz and Polosukhin, Illia},
  journal={Advances in neural information processing systems},
  volume={30},
  year={2017}
}

@article{var24,
  title={Visual autoregressive modeling: Scalable image generation via next-scale prediction},
  author={Tian, Keyu and Jiang, Yi and Yuan, Zehuan and Peng, Bingyue and Wang, Liwei},
  journal={Advances in neural information processing systems},
  volume={37},
  pages={84839--84865},
  year={2024}
}

@inproceedings{avg17,
  title={Communication-efficient learning of deep networks from decentralized data},
  author={McMahan, Brendan and Moore, Eider and Ramage, Daniel and Hampson, Seth and y Arcas, Blaise Aguera},
  booktitle={Artificial intelligence and statistics},
  pages={1273--1282},
  year={2017},
  organization={PMLR}
}

@article{dino,
  title={Dinov2: Learning robust visual features without supervision},
  author={Oquab, Maxime and Darcet, Timoth{\'e}e and Moutakanni, Th{\'e}o and Vo, Huy and Szafraniec, Marc and Khalidov, Vasil and Fernandez, Pierre and Haziza, Daniel and Massa, Francisco and El-Nouby, Alaaeldin and others},
  journal={arXiv preprint arXiv:2304.07193},
  year={2023}
}

@article{xray,
  title={Identifying medical diagnoses and treatable diseases by image-based deep learning},
  author={Kermany, Daniel S and Goldbaum, Michael and Cai, Wenjia and Valentim, Carolina CS and Liang, Huiying and Baxter, Sally L and McKeown, Alex and Yang, Ge and Wu, Xiaokang and Yan, Fangbing and others},
  journal={cell},
  volume={172},
  number={5},
  pages={1122--1131},
  year={2018},
  publisher={Elsevier}
}

@article{mpcpa24,
  title={MPCPA: Multi-Center Privacy Computing with Predictions Aggregation based on Denoising Diffusion Probabilistic Model},
  author={Luo, Guibo and Zhang, Hanwen and Wang, Xiuling and Chen, Mingzhi and Zhu, Yuesheng},
  journal={arXiv preprint arXiv:2403.07838},
  year={2024}
}

@inproceedings{officehome,
  title={Deep hashing network for unsupervised domain adaptation},
  author={Venkateswara, Hemanth and Eusebio, Jose and Chakraborty, Shayok and Panchanathan, Sethuraman},
  booktitle={Proceedings of the IEEE conference on computer vision and pattern recognition},
  pages={5018--5027},
  year={2017}
}

@inproceedings{pacs,
  title={Deeper, broader and artier domain generalization},
  author={Li, Da and Yang, Yongxin and Song, Yi-Zhe and Hospedales, Timothy M},
  booktitle={Proceedings of the IEEE international conference on computer vision},
  pages={5542--5550},
  year={2017}
}

@article{tiny,
  title={Lotteryfl: Personalized and communication-efficient federated learning with lottery ticket hypothesis on non-iid datasets},
  author={Li, Ang and Sun, Jingwei and Wang, Binghui and Duan, Lin and Li, Sicheng and Chen, Yiran and Li, Hai},
  journal={arXiv preprint arXiv:2008.03371},
  year={2020}
}

@misc{pmram,
  author       = {{orvile}},
  title        = {PMRAM: Bangladeshi Brain Cancer - MRI Dataset},
  year         = {2024},
  publisher    = {Kaggle},
  howpublished = {\url{https://www.kaggle.com/datasets/orvile/pmram-bangladeshi-brain-cancer-mri-dataset}},
  note         = {Accessed: 2025-07-31}
}

@article{mc,
  title={Lung segmentation in chest radiographs using anatomical atlases with nonrigid registration},
  author={Candemir, Sema and Jaeger, Stefan and Palaniappan, Kannappan and Musco, Jonathan P and Singh, Rahul K and Xue, Zhiyun and Karargyris, Alexandros and Antani, Sameer and Thoma, George and McDonald, Clement J},
  journal={IEEE transactions on medical imaging},
  volume={33},
  number={2},
  pages={577--590},
  year={2013},
  publisher={IEEE}
}

@article{sz,
  title={Two public chest X-ray datasets for computer-aided screening of pulmonary diseases},
  author={Jaeger, Stefan and Candemir, Sema and Antani, Sameer and W{\'a}ng, Y{\`\i}-Xi{\'a}ng J and Lu, Pu-Xuan and Thoma, George},
  journal={Quantitative imaging in medicine and surgery},
  volume={4},
  number={6},
  pages={475},
  year={2014}
}

@inproceedings{
vit,
title={An Image is Worth 16x16 Words: Transformers for Image Recognition at Scale},
author={Alexey Dosovitskiy and Lucas Beyer and Alexander Kolesnikov and Dirk Weissenborn and Xiaohua Zhai and Thomas Unterthiner and Mostafa Dehghani and Matthias Minderer and Georg Heigold and Sylvain Gelly and Jakob Uszkoreit and Neil Houlsby},
booktitle={International Conference on Learning Representations},
year={2021},
url={https://openreview.net/forum?id=YicbFdNTTy}
}

@article{fladvance21,
  title={Advances and open problems in federated learning},
  author={Kairouz, Peter and McMahan, H Brendan and Avent, Brendan and Bellet, Aur{\'e}lien and Bennis, Mehdi and Bhagoji, Arjun Nitin and Bonawitz, Kallista and Charles, Zachary and Cormode, Graham and Cummings, Rachel and others},
  journal={Foundations and trends{\textregistered} in machine learning},
  volume={14},
  number={1--2},
  pages={1--210},
  year={2021},
  publisher={Now Publishers, Inc.}
}

@inproceedings{pndm,
  author       = {Luping Liu and
                  Yi Ren and
                  Zhijie Lin and
                  Zhou Zhao},
  title        = {Pseudo Numerical Methods for Diffusion Models on Manifolds},
  booktitle    = {The Tenth International Conference on Learning Representations},
  publisher    = {OpenReview.net},
  year         = {2022},
  url          = {https://openreview.net/forum?id=PlKWVd2yBkY},
  bibsource    = {dblp computer science bibliography, https://dblp.org}
}

@article{ddim,
  title={Denoising diffusion implicit models},
  author={Song, Jiaming and Meng, Chenlin and Ermon, Stefano},
  journal={arXiv preprint arXiv:2010.02502},
  year={2020}
}


\end{document}